\documentclass[10pt,conference]{IEEEtran}
\usepackage{hyperref}
\usepackage[T1]{fontenc}
\usepackage{booktabs}
\usepackage{tabularx}

\usepackage{graphicx}
\graphicspath{{pngs/}}
\usepackage{color}

\IEEEoverridecommandlockouts
\begin{document}
\title{Integrating Anomaly Detection into Agentic AI for Proactive Risk Management in Human Activity}
\author{
\IEEEauthorblockN{Farbod Zorriassatine\thanks{Corresponding author: farbod.zorriassatine@gmail.com}}
\IEEEauthorblockA{
Department of Computer Science\\
Nottingham Trent University\\
Nottingham, UK\\
farbod.zorriassatine@ntu.ac.uk\\
ORCID: 0009-0001-4043-9603}
\and
\IEEEauthorblockN{Ahmad Lotfi}
\IEEEauthorblockA{
Department of Computer Science\\
Nottingham Trent University\\
Nottingham, UK\\
ahmad.lotfi@ntu.ac.uk\\
ORCID: 0000-0002-5139-6565}
}
\maketitle % typeset the header of the contribution
\begin{abstract}
Agentic AI, with goal-directed, proactive, and autonomous decision-making capabilities, offers a compelling opportunity to address movement-related risks in human activity, including the persistent hazard of falls among elderly populations. Despite numerous approaches to fall mitigation through fall prediction and detection, existing systems have not yet functioned as universal solutions across care pathways and safety-critical environments. This is largely due to limitations in consistently handling real-world complexity, particularly poor context awareness, high false alarm rates, environmental noise, and data scarcity. We argue that fall detection and fall prediction can usefully be formulated as anomaly detection problems and more effectively addressed through an agentic AI system. More broadly, this perspective enables the early identification of subtle deviations in movement patterns associated with increased risk, whether arising from age-related decline, fatigue, or environmental factors. While technical requirements for immediate deployment are beyond the scope of this paper, we propose a conceptual framework that highlights potential value. This framework promotes a well-orchestrated approach to risk management by dynamically selecting relevant tools and integrating them into adaptive decision-making workflows, rather than relying on static configurations tailored to narrowly defined scenarios.
% \keywords{Agentic AI, Anomaly Detection, Fall Detection, Fall Prediction, Gait}
\end{abstract}
% 150 Words
\section{Introduction}
There is currently extensive research focused on mitigating the impact of falls, particularly among the elderly people. Falls can result in serious injuries and, for those who survive, often lead to a significant decline in quality of life. They also impose severe financial burdens due to reduced independence and increased demand for healthcare services, including caregivers, treatment, and rehabilitation. One in three older adults experiences a fall, and the problem is worsening due to global population ageing driven by improved health and longevity.
Despite the development and even commercialisation of various sophisticated solutions—including the use of advanced Machine Learning (ML) and Artificial Intelligence (AI) techniques for fall detection and prediction—the rate of fall-related injuries has not declined. In fact, some reports suggest an increase in fall occurrences. While numerous technologies have attempted to address this pressing safety and welfare issue, no breakthrough has yet led to a widely adopted comprehensive solution \cite{dormosh2024systematic,gorce2025fall}. Existing approaches are often successful in specific contexts but fall short of offering a robust, consistently effective method for fall detection or prevention across care pathways. Persistent challenges, combined with limited trust in the explainability and reliability of alerting decisions, continue to hinder broader adoption and integration\cite{gorce2025fall}. As a result, the full potential of current fall mitigation technologies remains unrealised. This paper therefore explores how adopting Agentic AI (AAI) could help bridge these shortcomings.
\begin{figure}[t]
\centering 
\includegraphics[width=0.4\textwidth]{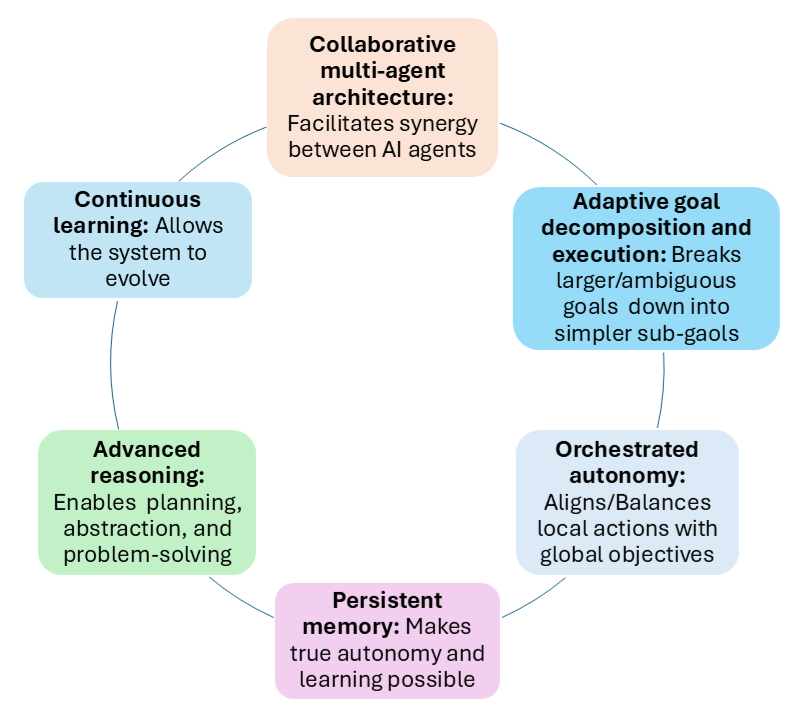}
\caption{The six core capabilities of an ideal Agentic AI system, shown as continuous operation rather than one-time setup.} 
\label{Fig1} % reqsAAI
\end{figure}
AAI is the latest development in advanced AI technology, based on the deployment of intelligent agents. However, to clarify the paper’s objectives, it is imperative to provide a precise definition of the intended AAI frameworks \cite{sapkota2025ai}, in order to distinguish them from the less advanced AI agents. Figure \ref{Fig1} provides a graphical representation of the core requirements for the intended AAI frameworks reflecting their dynamic, and iterative nature \cite{sapkota2025ai}. The most prominent advantage of AAI lies in its capability to operate autonomously in order to achieve predefined, yet often complex, goals. It is designed to perceive and interact with its environment, as well as with a wide array of components within that environment, to continuously learn and improve its own performance. Finally, it is also important to emphasise that Large Language Models (LLMs) often serve as a foundational component in implementing AAI's capabilities. LLMs are often combined in various ways to support complex multiple-step, dynamic reasoning and decision-making \cite{sapkota2025ai}. 

The rest of this paper is organised as follows: Section \ref{Fall-mitigation} highlights the existing work relevant to fall mitigation, thus, a concise summary of challenges and shortcomings is provided. In Section \ref{AD and falls} after providing a brief overview of Anomaly Detection (AD), it is argued how both FD and FP can be more efficiently and effectively addressed as variants of anomaly detection, applied exclusively in the domain of fall mitigation. Section \ref{Agentic_apparoach} introduces existing multi-agent systems for FM followed by AAI application for Anomaly Detection (AAI-AD). Section \ref{discussion} outlines strengths and challenges associated with the adoption of the transformative AAI technology and concludes that to provide a breakthrough in addressing falls adopting an agentic AI approach will be the most promising. 
\section{An Overview of Fall Mitigation}\label{Fall-mitigation} 
\begin{figure}[t]
\centering 
\includegraphics[width=0.5 \textwidth]{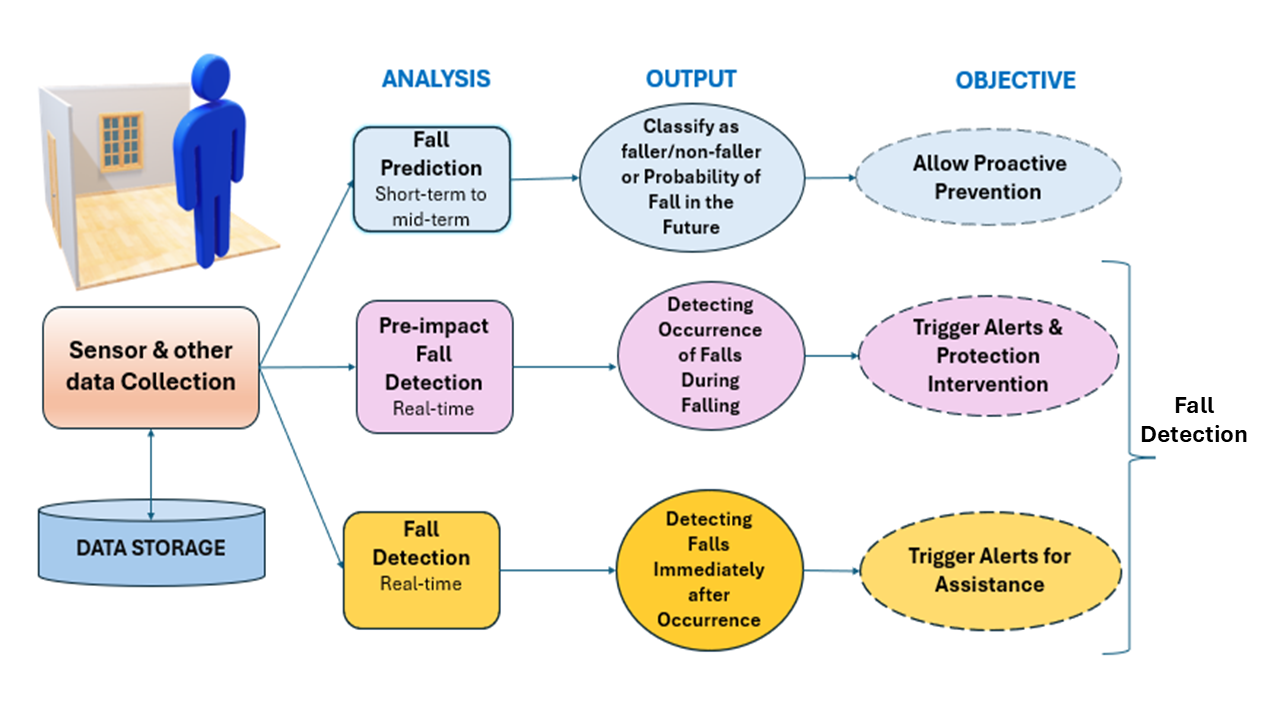}
\caption{An overview of the key groups of quantitative analytical methodologies commonly used in fall mitigation. Note: Fall risk assessment is excluded, as it primarily relies on qualitative and clinical inputs.} \label{fig2}
\end{figure}

The goal of Fall Mitigation (FM) is to reduce both the likelihood of falls and the severity of their consequences. Analytical FM methods are typically categorised into two distinct approaches: Fall Detection (FD) and Fall Prediction (FP) \cite{dormosh2024systematic,moyer2025artificial}. FD systems respond to fall events either immediately before or after they occur and are generally divided into two phases: pre-impact detection—identifying a fall during its onset \cite{ren2019research}, and post-event detection—recognising a fall after it has occurred \cite{gorce2025fall}. In contrast, FP systems aim to anticipate falls by assessing an individual’s risk level over a short-to-medium time frame (e.g., days to weeks or weeks to months) \cite{dormosh2024systematic}. Figure \ref{fig2} presents an overview of the key quantitative analytical methodologies for fall mitigation addressed in this paper, excluding fall risk assessment.  

While the terms FD and FP are sometimes used interchangeably, they represent fundamentally different concepts, as outlined in more detail below. Most of the existing literature has addressed FD and FP independently, rather than as integrated components of a unified FM framework. Comprehensive surveys and systematic reviews are available, offering detailed discussions of both approaches—see, for instance, \cite{dormosh2024systematic,moyer2025artificial} for FP and fall risk, and \cite{ren2019research} for FD techniques.

\subsection{Fall Detection Systems}\label{FD only}
FD systems can be broadly categorised into two distinct variants. The first, and most widely studied, is post-fall detection—commonly referred to simply as Fall Detection—which identifies a fall immediately after it has occurred \cite{gorce2025fall,ren2019research}. The second, less frequently addressed, is Pre-Impact Fall Detection, which aims to recognise a fall in progress—ideally milliseconds before the subject hits the ground—in order to activate preventative measures (e.g., airbags or robotic support) to reduce injury severity.

FD systems face several persistent challenges. These include: a lack of contextual awareness to allow more accurate and realistic judgement and understanding of the event; High rates of false positives and false negatives; A shortage of diverse and representative real-world datasets, particularly involving actual elderly people falls as opposed to simulated or Activities of Daily Living (ADL) data; low user acceptance and adoption; and privacy and ethical concerns surrounding the collection of personal and sensitive data \cite{gorce2025fall}.
\subsection{Fall Prediction Systems}\label{FP only}
Despite adopting a risk-based and proactive approach, FP systems also encounter significant challenges—many of which overlap with those faced by FD systems. These include imbalanced, limited, and biased datasets; environmental and sensor noise \cite{sczuka2023evaluating}; and issues with system reliability \cite{maruf2025state}.
However, FP introduces its own additional complexities. Unlike FD, which focuses on detecting the consequences of a fall (e.g., motion or posture anomalies), FP must account for the causes—which are multifactorial. This includes physiological, psychological, and behavioural risk factors, requiring sophisticated forecasting algorithms to model and interpret these diverse inputs \cite{subramaniam2022wearable}. As such, FP systems must handle broader data types, more intricate modelling, and deeper contextual understanding \cite{maruf2025state}.
Finally, it is important to distinguish Fall Risk Assessment (FRA)\cite{rango2025large} from FP. FRA is a broader concept that not only encompasses prediction but also includes long-term preventive strategies, ongoing risk monitoring, and deeper contextual analysis. While this paper deliberately excludes detailed discussion of FRA—including its potential framing as a long-term anomaly detection problem—it remains a crucial component of a comprehensive fall mitigation strategy. FRA will therefore be addressed in future work as part of the broader agentic AI framework.

\section{Anomaly Detection and Fall Mitigation}\label{AD and falls}
AD is a well-established technique for identifying rare yet critical events that are often difficult or infeasible to detect manually. It is especially valuable in contexts involving high-dimensional input spaces, large data volumes, and complex interdependencies. An anomaly, by definition, is any observation that deviates significantly from the norm—though this deviation can occur in many forms and to varying degrees \cite{samariya2023comprehensive}. 

Crucially, a wide range of problems across diverse fields—including healthcare, finance, cybersecurity, and industrial manufacturing—can be reframed as anomaly detection challenges. While the specific characteristics of anomalies differ by domain, the underlying principle of identifying unusual patterns remains consistent \cite{sunny2022anomaly}. This cross-domain translatability means that many existing, mature methodologies from AD—originally developed for domains like cybersecurity can often be adapted to new problem spaces without reinventing the wheel.

The field of anomaly detection is supported by a rich and well-established body of knowledge. This includes definitions that delineate the scope of anomalies (i.e. three main types of anomalies: point, contextual, and collective anomalies). There is also an extensive set of common methodologies (including tools and frameworks (e.g. statistical, threshold-based rules, machine learning, and deep learning techniques). Shared challenges including limitations, open problems, and obstacles (e.g. concept drift, data imbalance) are also extensively documented. For the future directions comprising emerging trends, and research opportunities see extensive reviews such as \cite{samariya2023comprehensive} and \cite{belay2026agentic}.
Extending the above argument to fall mitigation, many commonly discussed fall-related problems can be reframed as anomaly detection issues rather than being expressed using FM-specific language. For example, this includes using anomaly scoring instead of fall risk assessment, early anomaly detection instead of pre-fall activity detection, and concept drift instead of progressive decline in mobility. More specifically, the two constituent components of FM can be expressed in AD terms as follows: FD corresponds to real-time point-anomaly detection of abrupt posture changes (e.g. see \cite{ren2019research}), while FP corresponds to subtle-change or early-warning anomaly detection (among many closely related alternatives) \cite{kocuvan2023predicting}. Finally, FM frameworks that treat falls as context-dependent rather than purely motion-based can be formulated as contextual anomaly detection problems (e.g. see \cite{samani2022anomaly}). In the following section, we show how this connection between AD and FM can be exploited to adapt existing work on agentic AI for anomaly detection (AAI-AD) and accelerate progress in this area.

\section{The Vision: Agentic AI Applied to Fall Mitigation}\label{Agentic_apparoach}
A growing body of experimental research on fall mitigation has explored the 
use of Multi-Agent Systems (MAS) to distribute tasks among intelligent AI 
agents, focusing on fall detection \cite{pillai2024improving}, fall risk prediction \cite{rango2025large}, or broader 
aspects of elderly care \cite{khan2024adaptive}. Earlier conceptual work  \cite{kaluvza2011multi} proposed 
sophisticated monitoring systems but remained largely theoretical, whilst more 
recent efforts\cite{wen2025evolution} have implemented reactive multi-agent systems using ML for specific tasks.

However, the agents proposed across this literature range from simple, 
rule-based models to more complex, task-specific implementations that fall 
short of true AAI. Many studies use the term "autonomous" loosely, referring 
to narrowly automated systems \cite{ramanujam2022aifms}, and even those claiming proactivity or adaptivity \cite{ahamad2024omobot} often lack the two-way interactivity required to enable multiple specialised agents to collaboratively pursue complex objectives 
through structured communication and shared contexts \cite{sapkota2025ai}. While a few works \cite{pillai2024improving,rango2025large} show closer alignment with AAI principles, most do not yet support genuine agentic capabilities. Some adopt scopes so broad that fall mitigation becomes merely one of many general "health incidents" \cite{khan2024adaptive}.

This highlights a clear gap: the need for conceptual and design guidance 
toward truly agentic fall mitigation systems.
\subsection{Agentic AI for Anomaly Detection (AAI-AD)}\label{AAIAD} 
There are several reasons why Anomaly Detection (AD) aligns well with AAI, 
including its autonomy (e.g. unsupervised learning without labelled data), 
perceptiveness (learning normal patterns across modalities), and context-
awareness. AD's continual search for anomalies also supports AAI's proactive, 
self-improving nature \cite{timms2024agentic}. However, traditional AD systems operate largely in isolation, lacking the goal-oriented reasoning and dynamic planning that 
characterize true AAI. By integrating AD within an agentic framework, anomaly 
management can be transformed from static, human-in-the-loop processes into 
autonomous, goal-driven intelligence capable of reasoning, adapting, and 
acting effectively under real-world complexity \cite{barenji2025agentic}. These properties make AD well-suited for such transformation through AAI, enabling adaptability to 
evolving data, decentralised scalability via multi-agent systems, and improved 
reasoning with LLM-based agents \cite{belay2026agentic}.

Despite growing interest, few AI-agent AD systems fully meet the defined 
attributes of AAI. For instance, authors in \cite{yang2023adt} addressed static thresholds 
in AD using DQNs and autoencoders, but lacked generative reasoning to form 
new logical inferences or create novel insights beyond anomaly scoring—hence 
not fully AAI-compliant. Conversely, authors in \cite{park2024enhancing} introduced a collaborative LLM-based multi-agent framework for financial anomaly detection, reducing reliance on human intervention. A comprehensive review of recent advancements at the intersection of agentic AI and multimodal anomaly detection is provided in \cite{belay2026agentic}.

Building on the argument presented in Section  \ref{AD and falls} that FM is a form of AD, 
and drawing on workflow templates from models such as AD-Agent \cite{yang2025ad} - which incorporate components including task parsing, model selection, knowledge retrieval, code synthesis and debugging, performance assessment and hyperparameter tuning - we propose a comprehensive AAI framework specifically designed for fall mitigation: the AD-based Fall Mitigation Agentic AI System (ADFM-AAI). This framework, outlined in the following section, demonstrates how AAI principles can be systematically applied to address the complex, multi-faceted challenge of fall mitigation through anomaly detection.

\subsection{Proposed Conceptual Framework and Architecture of ADFM-AAI}
\label{conceptual framework}
Here, we briefly outline our proposed AAI concept and framework and how it maps to its key components.
% \subsubsection{Core Agentic Capabilities:}
% \label{Core-ag-caps}
\subsubsection{Mapping the Requirements of AD for FM to Core Agentic Capabilities}\label{mapping}
ADFM-AAI must determine how complex AD tasks related to fall mitigation can be performed proactively and autonomously, with minimal human input, particularly on a continuous basis. While the six core principles of Agentic AI (see Figure \ref{Fig1}) underpin the proposed ADFM-AAI framework, these can be synthesised into four consolidated core capabilities for conceptual clarity and functional cohesion:
\begin{itemize}
 \item \textbf{Perception \& Data Acquisition} (i.e. receive comprehensive and diverse data streams related to gait, ADL, and the environment through various types of sensors including wearable, ambient, and vision),
 \item \textbf{AD \& Reasoning} (i.e. analyse data, and determine the nature of events in real-time or for short to long term, especially for identifying anomalous conditions - i.e. point: such as a fall after a person is lying on the floor, contextual: e.g. loss of balance or near-fall before actual fall, or collective or subtle-change: early signs of upcoming falls e.g. detecting gait instability, or a sequence of unusual ADL events),
 \item \textbf{Agentic Orchestration \& Planning} (i.e. decide the next steps and courses of action to take; e.g. recommend personalised training or preventive adjustments to the environment if a fall is predicted to happen in the near future, plan to trigger pre-impact remedies such as raising guard-rails or air-bags if imminent fall is predicted, call the emergency services if a fall has occurred, or identify which external tools to engage such as medication databases, or sensor networks),
 \item \textbf{Action, Feedback \& Learning} (i.e. implement and execute the steps planned by the preceding core capability). 
\end{itemize}

\begin{figure}
    \centering
    \includegraphics[width=1\linewidth]{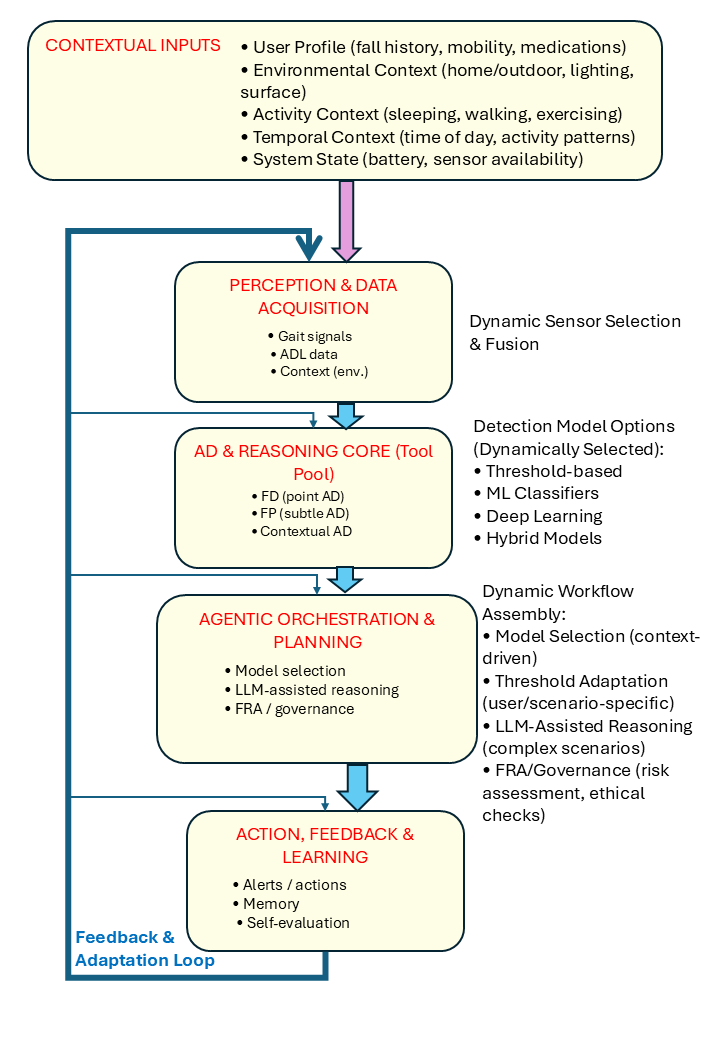}
    \caption{High-level architecture of the proposed AD-based Fall Mitigation Agentic AI system (ADFM-AAI), illustrating the minimal set of mission-critical components}
    \label{Fig3}
\end{figure}
 Figure \ref{Fig3} presents a streamlined view of the proposed ADFM-AAI, organised around a minimal set of mission-critical components. Data capturing Gait, Activities of Daily Living (ADL), and environmental conditions are continuously collected and processed through the Perception and Data Acquisition Layer. This processed information serves as input for the AD-Centred Reasoning Core, which identifies three categories of gait-related anomalies: point, collective (subtle-change), and contextual. Depending on the identified signal patterns, these anomalies may indicate a fall risk manifest across different timescales—ranging from long-term trends and medium-term risks to immediate or post-occurrence events. The system’s `brain'—the Orchestration and Planning Core—evaluates these findings to determine the most suitable configuration of anomaly detection and fall mitigation models. By analysing the dynamic situation, this core can adapt individual components or the entire system to anticipate upcoming risks. Consequently, the most appropriate decisions are formulated and forwarded to the Action, Feedback, and Learning Layer, which incorporates continuous feedback, memory, and self-evaluation. The resulting architecture provides a closed, goal-directed loop where dedicated agents retrieve personalised gait histories, anomaly patterns, and relevant clinical, environmental, and knowledge-base information (both structured and unstructured) to provide a continuous, adaptive anomaly-detection process for fall mitigation. The roles of agents and LLMs within the proposed fall mitigation architecture are further described below.

\subsubsection{Agents}\label{sub:agents}
Agents embody specific tasks, such as those involved in fall management (e.g., FD). Within the FM framework, agents are defined according to related tasks and sub-tasks, and may be grouped by criteria such as function, behaviour, or architectural layer. Given the potentially large number of useful agents, only the most critical examples relevant to FM are presented in Table \ref{tab:LLMs_agents}.

\subsubsection{LLM as a Central/Shared Reasoning Service}\label{sub: LLM}
LLMs and some of the agents presented in Table \ref{tab:LLMs_agents} can collaborate to form an intelligent, task-oriented FM system. LLMs operate exclusively on textual data, supplied by agents through appropriate translation or formatting. Different types of LLMs may be developed based on operational and task specific requirements. These LLMs include lightweight, large, domain-specific, instruction tuned, tool-augmented, and multi-modal types \cite{rango2025large,sapkota2025ai}. 
LLMs provide a range of cognitive capabilities for the agents or other involved components: e.g. interpreting caregiver's instruction, reasoning that a person is unstable after observing several slips in recent days, planning more frequent gait monitoring when suspecting recent gait instability, making decision to alert care givers for more vigilance, or recommending to the clinicians for offering corrective training. LLMs also enable two-way communication; between different agents, between multiple LLMs, between different modules of ADFM-AAI and human users (e.g. residents, patients, or caregivers), or between agents and LLMs—offering the distinctive advantage of the AAI approach to functional modelling over conventional AI solutions.

\begin{table}[bt!]
 \centering
 \caption{Listing of Potential AI Agents for inclusion in ADFM-AAI} 
 \label{tab:LLMs_agents}
  % This reduces the font size specifically for this table
 \scriptsize 
  % Reducing column padding to make it even more compact
 \setlength{\tabcolsep}{2pt} 
  \begin{tabular}{| p{2.5cm} | p{6cm} |}
 \hline
 \textbf{Layer} & \textbf{Agent Category} \\ 
 \hline
 Data Gathering & Data Preprocessing Agents, Ingestion Agents \\
 \hline
 Analysis \& Prediction & Intelligent Feature Engineering, FM, Anomaly Detection \\
 \hline
 Action \& Intervention & Triggering proactive, non-urgent interventions, executing emergency protocols, Alerting Agents, Interacting with IoT devices \\
 \hline
 Orchestration & Agent supervision, Explainability (XAI), Human-Agent Collaboration, Governance and Safety \\
 \hline
 Fall Mitigation & FD, FP, pre-impact FD, FRA \\
 \hline
 AD & AD-Processor, AD-model-Selector, AD-Info Miner, AD-Code Generator, AD-Reviewer, AD-Evaluator, AD-Optimiser, concept-drift, data-imbalance \\
 \hline
 \end{tabular}
\end{table}
\section{Discussion}\label{discussion}
As discussed earlier, FM can be viewed through AD. However, traditional AD approaches often prove to be inflexible and fragile when confronted with the wide range of anomaly types, diverse data modalities, high data volumes, and the dynamic nature of real-world environments—challenges that are equally evident in the FM domain.

Our paper outlines a holistic vision to reduce falls among older adults, urging stakeholders to move beyond fragmented FD and FP methods toward a more unified FM framework. Conventional approaches often fail due to a static, limited view of the problem space. In contrast, real-world scenarios demand an adaptive approach—one that can dynamically select and deploy the most effective combination of solutions from a broad array of FM options, including various sensors, algorithms, and data fusion techniques.

In this context, the autonomy and reasoning capabilities of the proposed ADFM-AAI offer a more promising alternative. By continuously adapting its strategy in real time, ADFM-AAI can address the complexities of individual cases and their environments more effectively than any siloed approach deployed to date. 

While intelligent approaches to AD in gait and ADLs are advancing, none yet exhibit true agency or goal-driven reasoning. Adopting AAI is complex, involving layered implementation and heightened risks—especially where autonomy in fall mitigation could pose life-threatening consequences.

Though AAI shows strong potential for comprehensive solutions, its integration will be gradual. With sufficient investment and collective effort, it is possible to envision a future where a robust infrastructure exists to protect all individuals, especially the elderly people from the devastating consequences of falls, which currently affect one-third of the older adult population.

\section*{Acknowledgement}
Generative AI tools were used to assist with language editing and clarity. All technical content, analysis, and conclusions
remain the responsibility of the authors.

\bibliographystyle{splncs04}
% \bibliography{myref}

\end{document}